\documentclass[10pt,twoside,reqno,final]{manuscript}

\usepackage[numbers,sort&compress]{natbib}

\usepackage[notcite,notref]{showkeys}

\hypersetup{
  colorlinks=true,
  linkcolor=magenta,
  citecolor=blue,
  filecolor=blue,
  urlcolor=blue
}

\graphicspath{{./figures/}}

\newcommand{\abs}[1]{\lvert#1\rvert}

\title{Solving Rubik’s Cube Without Tricky Sampling}

\shorttitle{Solving Rubik’s Cube Without Tricky Sampling}

\author[1,\cormark]{Yicheng Lin}
\author[2]{Siyu Liang}
\affil[1]{School Of Basic Medical Sciences, Fudan University}
\affil[2]{School Of Mechanical Engineering, Tongji University}
\cortext{Author e-mail addresses:  \texttt{linyc20@fudan.edu.cn, cedric$\_$liang@163.com}}

\shortauthor{Yicheng L and Siyu L}

\begin{document}

\maketitle
\thispagestyle{firststyle}

\begin{abstract}
The Rubik’s Cube, with its vast state space and sparse reward structure, presents a significant challenge for reinforcement learning (RL) due to the difficulty of reaching rewarded states. Previous research addressed this by propagating cost-to-go estimates from the solved state and incorporating search techniques. These approaches differ from human strategies that start from fully scrambled cubes, which can be tricky for solving a general sparse-reward problem. In this paper, we introduce a novel RL algorithm using policy gradient methods to solve the Rubik’s Cube without relying on near solved-state sampling. Our approach employs a neural network to predict cost patterns between states, allowing the agent to learn directly from scrambled states. Our method was tested on the 2x2x2 Rubik’s Cube, where the cube was scrambled 50,000 times, and the model successfully solved it in over 99.4\% of cases. Notably, this result was achieved using only the policy network without relying on tree search as in previous methods, demonstrating its effectiveness and potential for broader applications in sparse-reward problems.


\end{abstract}


\section{Introduction}

The Rubik’s Cube is a classic combinatorial puzzle that presents unique and significant challenges for artificial intelligence and machine learning \cite{ref_1, ref_2, ref_3}. With an enormous number of possible states, only a single state represents the solved configuration, making it particularly difficult for reinforcement learning algorithms to tackle. The challenge lies in the fact that, during training, the moves taken by the agent’s policy are exceedingly unlikely to reach the solved state. This lack of reinforcement for long stretches of exploration often leaves the agent struggling to discover any rewarded state, making it difficult to learn an effective strategy.

This highlights a key issue in sparse reward problems within reinforcement learning, where the absence of frequent feedback severely hinders the agent’s ability to learn \cite{ref_4,ref_5,ref_6}. Without enough guidance, it becomes much harder for the agent to develop useful policies. The Rubik’s Cube serves as an ideal testbed for addressing these challenges, and developing reinforcement learning algorithms capable of solving this puzzle could offer broader insights into how to handle sparse-reward environments in other domains.

While classical methods for solving the Rubik’s Cube have existed for decades \cite{ref_7}, it wasn’t until the work of Agostinelli et al. \cite{ref_3}, who developed the DeepCube algorithm, that an AI system using reinforcement learning \cite{ref_8,ref_9,ref_10,ref_11,ref_12,ref_13} was able to solve the Rubik’s Cube from any starting configuration. In their follow-up work, DeepCubeA \cite{ref_2}, they implement Deep Approximate Value Iteration (DAVI), which iteratively estimates the cost-to-go for states near the solved state and propagates this information outward to states further away. This process involves scrambling a solved cube multiple times and collecting trajectories that start from the solved configuration. While this approach allows DeepCube to effectively solve the Rubik’s Cube, it contrasts with the way humans typically solve the puzzle. Humans start with a fully scrambled cube, without prior knowledge of states close to the solved state. We observe states far from the solution and gradually develop an understanding of the underlying structure and relationships between states, even when they are distant from the solved state.

In this paper, we address this issue by developing a new reinforcement learning approach that uses policy gradient methods to solve the Rubik’s Cube without sampling states from a solved configuration. Instead, we continuously sample states from a fully scrambled cube and build up rewards based on the underlying distance patterns between states. Unlike methods mentioned above, which rely on search methods like Monte Carlo Tree Search (MCTS), our approach requires no such search techniques. Using this method, we successfully solved the 2x2x2 Rubik’s Cube with a success rate of 99.4\% across 50,000 test cases.

\section{Methods}

\subsection{State and action spaces}
\begin{figure}[htpb]
  \centering
  \includegraphics[width=0.7\linewidth]{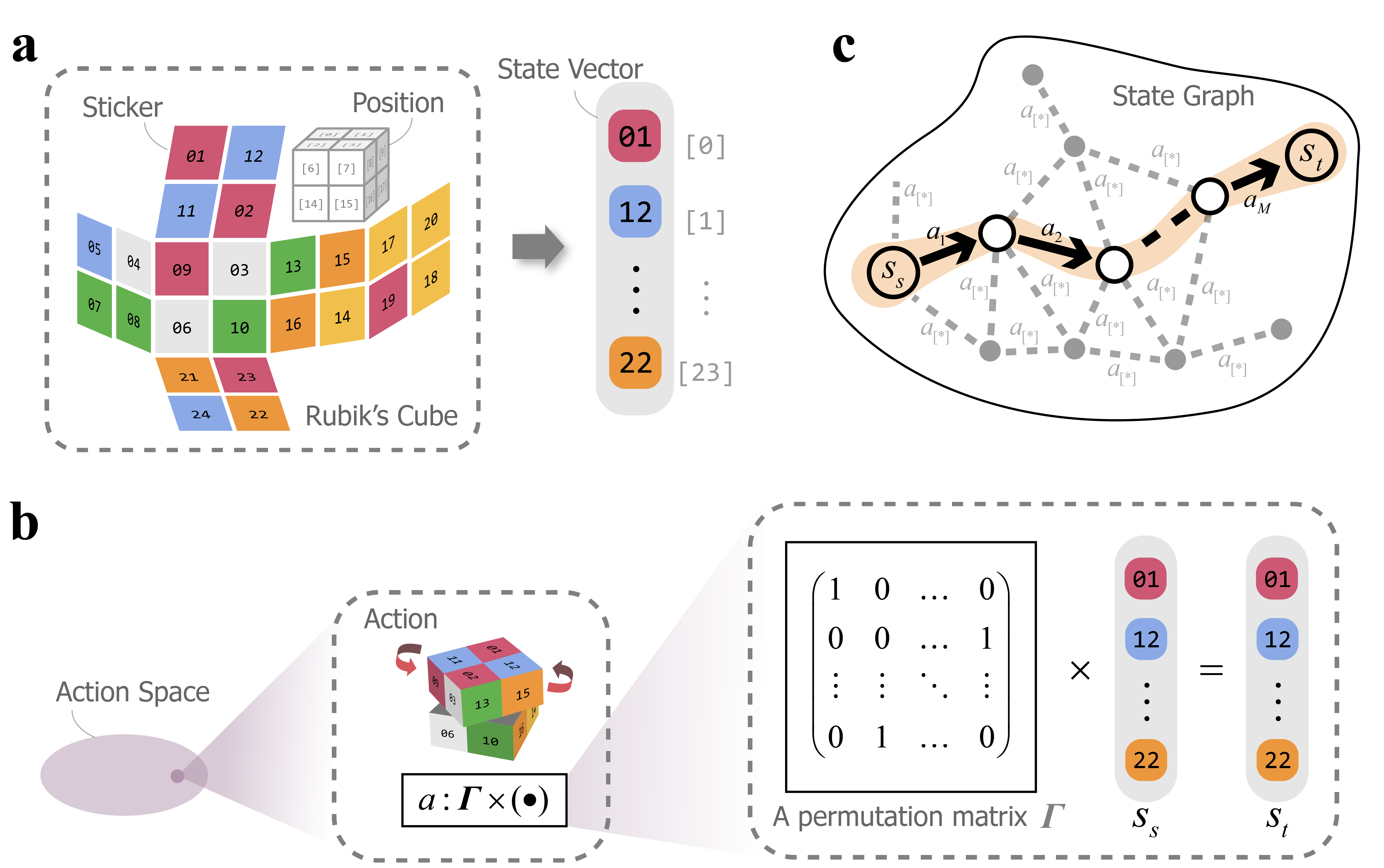}
  \caption{General Representation of Rubik's cube problem. a. The state of the cube can be represented by a vector, with each sticker encoded as a number. b. Scrambles are modeled using a permutation matrix applied to the state vector. c. The cube’s topological structure is visualized as a state graph, where nodes represent states and edges represent transitions.}
  \label{fig_1}
\end{figure}

A Rubik’s Cube consists of a number of stickers, each uniquely associated with a specific position on the cube.
In general, for any given Rubik’s Cube, all stickers can be encoded as elements of a set of numbers.
The state of the Rubik’s Cube is then defined as a vector, where each element corresponds to the encoded value of a sticker, recorded in a specific order (Figure~\ref{fig_1}a).
This vector is denoted as $s$, with its dimension $N$ representing the total number of stickers.
An action performed on a Rubik’s Cube involves scrambling the cube hence rearranging specific stickers according to a predefined rule.
Using the state vector representation, an action $a$ applied to a given state $s$ can be defined as a function $a\left(s\right)=\mathbf{\Gamma}s$, where $\mathbf{\Gamma}$ is a permutation matrix of size $N\times N$, representing the specified rearrangement rule (Figure~\ref{fig_1}b).
The action space $\mathcal{H}$ of a Rubik’s Cube is defined as the set of all possible actions, each determined by a specific rearrangement rule.
Applying an action $a_{\left[\ast\right]}$ in$\mathcal{H}$ to a given state $s_s$ results in a new state $s_t=a_{\left[\ast\right]}(s_s)$.
Starting from an initial state $s_0$, repeated applications of actions will cause the Rubik’s Cube to transition through a sequence of states.
The set of all states that can be reached from $s_0$ is defined as the state space derived from $s_0$, denoted as $S$.
Consequently, the topological structure of the Rubik’s Cube can be represented as a state graph, where each node corresponds to a state, and edges represent actions connecting these states.
Solving the Rubik’s Cube can thus be equivalently formulated as a pathfinding problem in the state graph: given a pair of states $(s_s,\ s_t)$, the goal is to identify a feasible sequence of actions $\left\langle a_i\right\rangle_M$ that transforms $s_s$ into $s_t$, with $s_t=a_M(\ldots a_i(\ldots a_2(a_1(s_s))))$, where $a_i\in A$,$i=1,2,\ldots,M$, with $M$ the length of the action sequence (Figure~\ref{fig_1}c).

\begin{figure}[htpb]
  \centering
  \includegraphics[width=0.7\linewidth]{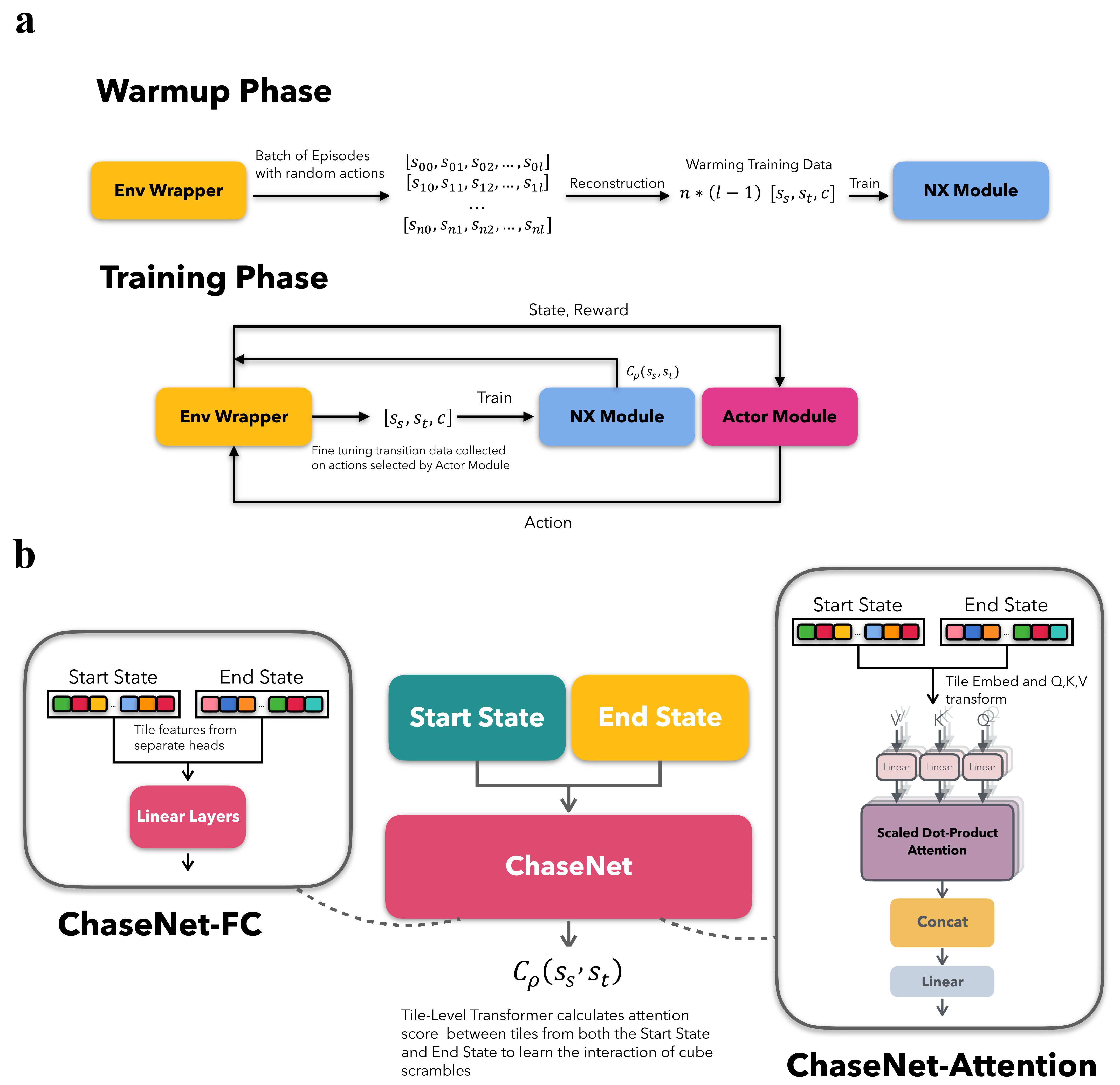}
  \caption{The NX Module a. The NX Module training process is divided into a warmup phase and a training phase. b. Two architectural variants of ChaseNet: ChaseNet-FC (fully connected) and ChaseNet-Attention (attention-based).}
  \label{fig_2}
\end{figure}

\subsection{Cost of states pairs}
We define the cost between state pair $(s_s,s_t)$ as the minimal number of scrambles needed from the start state $s_s$ to the target state $s_t$. It can also be represented as the length of the optimal path which linked the state pair $(s_s,s_t)$ in the state graph.

\subsection{NX Module}
The core objective of the NX Module is to collect states far from the solved state and train a model to accurately estimate the cost between any pair of states based on their sticker representation.
In this module, a neural network $C_\rho(s_s,s_t)$ termed ChaseNet, parameterized by $\rho$, was trained to estimate the cost between pairs of states $(s_s,s_t)$ using samples generated from scrambled cubes. During the warmup phase, state pairs $(s_s,s_t)$ are created by applying random actions sampled from the action space, with the restriction that no action is repeated more than three times in a row, as this would return the cube to a prior state, producing incorrect labels. We emphasize that the $s_s$ collected in the trajectories are randomly scrambled, without any restriction on their distance from the solved $s_g$, unlike in previous works. In the training phase, state pairs are generated by applying actions sampled from the policy network $p_\theta(a|s)$, allowing the model to refine its predictions based on learned policies.

\begin{algorithm}
\caption{Warmup for NX-Module. \textbf{Input}: $B$: Dataset Size, $K$: Maximum number of twists, $J$: training iterations. 
\textbf{Output}: $\rho$: the trained ChaseNet parameters.}
\label{alg_1}
\begin{algorithmic}
    \State {$\rho \gets $INITIALIZENETWORKPARAMETERS}
    \For{$j=1,2,\dots,J$ }
        \State {INITIALIZEDATASET $D\gets \emptyset$}
        \While{$\abs{D}<B$}
            \State{Generate a random initial scrambled state $s_s$}
            \State{$s_{current}\gets s_s$}\Comment{Set current satate}
            \For{$i=1,2,\dots,K$}
                \State{Apply a random scramble to get the next state $s_i$}
                \State{Compute the cost $y_i=i$ (the number of twists applied)}
                \State{$D\gets\ D\cup{(s_s,s_i,y_i)}$}\Comment{Update dataset}
                \State{$s_{current}\gets s_i$}\Comment{Update current state} 
            \EndFor
        \EndWhile
        \State{$\rho\gets \mathrm{Train}(C_\rho,X,y)$ with each $X_i=(s_s,s_i)$ in $D$}
        
    \EndFor
\end{algorithmic}
\end{algorithm}

\subsection{Env Module}
The environment is wrapped into the Env Module. During interaction with the agent, it takes in the action $a_i$ and returns the consequent state $s_{i+1}$ and the reward $r_i$, with $i$ the index within the episode. The reward is formulated by the following:

\begin{equation}
    r_i=-\log_b{C_\rho(s_{i+1},s_g)}
\end{equation}

where $s_g$ is the resolved state and $b=1.2$ is a base specified to rescale the direct predicted cost of state pair $(s_s,s_g)$. Specifically, when the goal state is reached, a fixed reward of 100 is assigned.

\subsection{Actor Module}
The Actor Module refines the policy network $p_\theta\left(a\middle| s\right)$ with Proximal Policy Optimization (PPO) \cite{ref_14} algorithm. Specifically, given the reward returned from the Env Module which is formulated using the cost of the current state to the solved state, the objective function is formulated as following:

\begin{equation}
    L\left(s,a,\theta_k,\theta\right)=min\left(\frac{p_\theta(a\mid s)}{p_{\theta_k}(a\mid s)}A^{p_{\theta_k}}(s,a),{\ }g\left(\epsilon,{\ A}^{p_{\theta_k}}(s,a)\right)\right)
    \label{eq_L1}
\end{equation}

where

\begin{equation}
    g\left(\epsilon,A\right)=
    \left\{\begin{matrix}
    (1+\epsilon) A\quad A\ge 0 \\
    (1-\epsilon) A\quad A<0
\end{matrix}\right.
\label{eq_L2}
\end{equation}

The advantage $A^{p_\theta}\left(a\right|s)$ is the difference between the Q-value $Q^{p_\theta}(s,a)$, the expected return of selection action a in state s, and the value $V^{\varphi_k}(s)$, the predicted return of state s by the critic network parameterized by $\varphi$ in the $k$-th iteration. $\epsilon$ is a hyperparameter that we set at 0.2, as used in the previous paper \cite{ref_14}.

\begin{algorithm}
\caption{Training for Actor Module and NX module finetune. \textbf{Input}: $C_\rho$: Trained ChaseNet model parameterized by $\rho$, $p_\theta$: Policy network parameterized by $\theta$, $J$:Training iterations, $B$: Batch size, $\epsilon$: PPO clipping parameter, $K$: Number of twists. 
\textbf{Output}: $\rho$: The trained ChaseNet parameters, $\theta$: The trained policy network parameters.
}
\label{alg_2}
\begin{algorithmic}
    \For{$j=1,2,\dots,J$}
        \State{Generate a random initial scrambled state $s_s$}
        \State{$s_{current}\gets s_s$}\Comment{Set current state}
        \State{$B_f \gets \emptyset$}\Comment{Initialize storage buffer for states, actions, rewards and log probabilities}
        \State{$D\gets\emptyset$} \Comment{Initialize Dataset $D$ for ChaseNet fine tuning}
        \State{$i\gets 0$} \Comment{Index of states in an episode}
        \While{Episode not end}
            \State{Select action $a_i~p_\theta(a|s_i)$}
            \State{Apply action $a_i$ to get next state $s_{i+1}$}
            \State{$r_i\gets-\log_b{C_\rho(s_{i+1},s_g)}$} \Comment{Predict cost using ChaseNet with $s_g$ the resolved state}
            \State{Store $\left(s_i,a_i,r_i,\log p_\theta\left(a_i\middle| s_i\right)\right)$ in $B_f$}
            \State{$s_i\gets s_{i+1}$}\Comment{Update state}
            \State{$D\gets\ D\cup{(s_s,s_i,i)}$}\Comment{Collect Data for ChaseNet fine tuning}
            \State{$\rho\gets FINETUNE(C_\rho,X,y)$ with each $X_i=(s_s,s_i)$ in $D$}
            \If{$\abs{B_f}>B$ }
                \State{Compute objective $L_{policy}\left(s,a,\theta_k,\theta\right)$ using formula \ref{eq_L1} with $s$, $a$ sampled from $B_f$}
                \State{Update Policy network parameters $\theta$ by maximizing $L_{policy}$}
            \EndIf
        \State{$i\gets i+1$}
        \EndWhile   
    \EndFor
\end{algorithmic}
\end{algorithm}

\subsection{Network Architectures}
For our comparative study, we developed ChaseNet to predict the cost between state pairs, implementing two distinct neural network architectures: ChaseNet-FC, composed entirely of fully connected layers, and ChaseNet-Attention, which utilizes a sticker-level transformer \cite{ref_15} architecture, as described below.
ChaseNet-FC offers a straightforward architecture, with two linear heads that independently extract features from the flattened embedding vectors of the start state $s_s$ and end state $s_t$. These features are then concatenated and passed through additional linear layers to produce the final output cost. 

ChaseNet-Attention used a sticker-level transformer that computes attention scores between stickers from both the start state $s_s$ and end state $s_t$. This enables the model to learn the interactions between cube scrambles more effectively. Specifically, batches of vector representations of $s_s$ and $s_t$ are combined to form the input tensor $X$, which is of shape $B\times M$, where $B$ denotes the batch size and $M=2\times N$ with $N$ the total number of stickers on the cube. The attention scores $A$ are computed between each position in the combined input tensor $X$:

\begin{equation}
    A_{ij}=Attention\left(X_i,X_j\right),\ \forall i,j\in[1,2,\ldots,M]
\end{equation}

where $X_i$ and $X_j$ correspond to the $i$-th and $j$-th positions in the combined input tensor. This approach allows the model to capture the relationships and dependencies between every sticker’s position in the start state and end state. By leveraging these attention scores and the transformer’s ability to capture intricate dependencies between sticker positions, ChaseNet-Attention can possibly learn to predict the cost between states more effectively than the approach of ChaseNet-FC. 

In the Actor Module, the policy network $p_\theta\left(a\middle| s\right)$ consists of linear layers that extract features from a single state $s$ and output a probability distribution over the discrete action space for the current state.

\section{Results}

\subsection{Comparison of performance of ChaseNet-FC and ChaseNet-Attention in the warmup phase}

\begin{figure}[htpb]
  \centering
  \includegraphics[width=\linewidth]{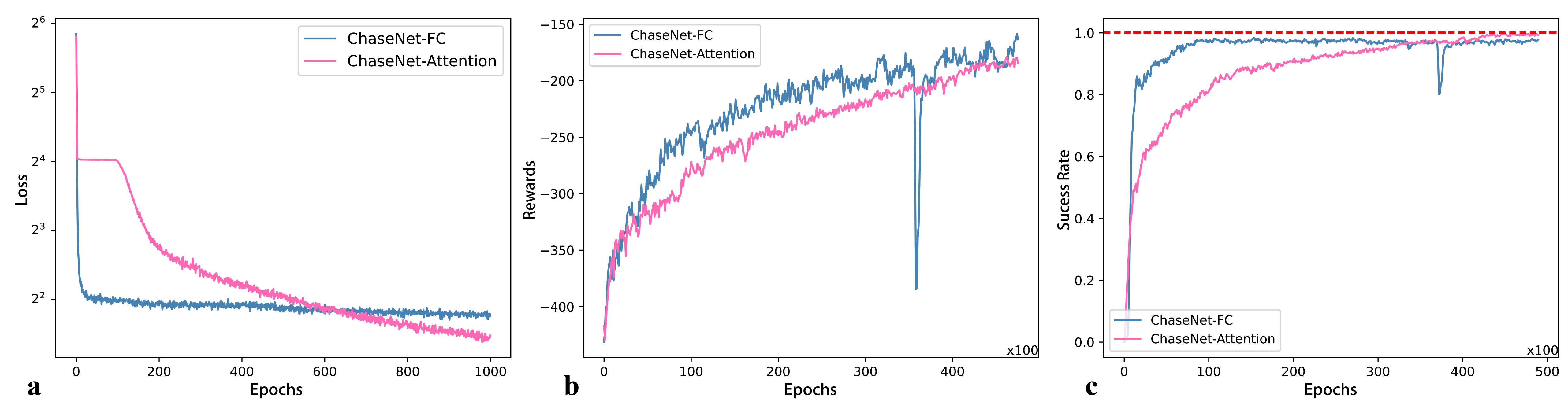}
  \caption{a. Warmup loss for ChaseNet-FC and ChaseNet-Attention. b. Average rewards during RL training using rewards from ChaseNet-FC and ChaseNet-Attention. c. Success rate during RL training using rewards from ChaseNet-FC and ChaseNet-Attention.}
  \label{fig_3}
\end{figure}

In this work, we focus on the 2x2x2 Rubik’s Cube as a challenging yet computationally feasible problem for evaluating our approach. We monitored the loss curves of both ChaseNet-FC and ChaseNet-Attention across multiple epochs. With the training iteration set to $J=1000$, the resulting loss curves are displayed in Figure~\ref{fig_3}a, providing a comparative view of the models’ convergence behavior.
ChaseNet-FC demonstrates a faster convergence rate than ChaseNet-Attention, likely due to its simpler architecture composed solely of linear layers, which enables more straightforward feature extraction and quicker optimization. In contrast, ChaseNet-Attention, which incorporates a sticker-level transformer for capturing more complex spatial dependencies, initially converges more slowly. However, after 1,000 iterations of warm-up training, ChaseNet-Attention achieves a lower final loss value, suggesting that its architecture, though more complex, ultimately captures richer state representations that improve prediction accuracy. We evaluated both ChaseNet-FC and ChaseNet-Attention on an independent test set to measure Spearman’s correlation coefficient between their outputs and the true cost values. ChaseNet-FC achieved a coefficient of 0.834, while ChaseNet-Attention reached a higher coefficient of 0.901, indicating that the sticker-level transformer in ChaseNet-Attention provides a more accurate prediction of state-pair costs. 

\subsection{Solving 2x2x2 Rubik’s Cube without searching}

After the warm-up phase of training ChaseNet, we used its predictions to train the policy network via the Proximal Policy Optimization (PPO) algorithm. For comparative analysis, we evaluated the performance of the policy network trained using rewards from both ChaseNet-FC and ChaseNet-Attention. During testing, we scrambled the cube and measured the number of times the policy network successfully solved it. Notably, this testing phase involved no tree search; instead, we relied solely on the policy network to guide each move. Despite the absence of search, the policy network achieved a remarkable success rate. Here, the success rate is defined as the ratio of successful solves to the total number of 50 test attempts. Figure~\ref{fig_3}b-c shows the performance of both ChaseNet-FC and ChaseNet-Attention. During the final validation, we achieved a success rate of over 99.4\% across 50,000 test cases.

\section{Discussion}
In this work, we presented a reinforcement learning approach for solving the Rubik’s Cube without relying on sampling starting from solved states. Unlike methods like DeepCubeA, which sample states near the solution, our approach learns a solution policy directly from learning the cost patterns of fully scrambled states. By leveraging ChaseNet to accurately estimate state transition costs, the NX Module guides policy optimization effectively from random starting points, aligning with the way humans solve the puzzle by building heuristics from disordered states. Achieving a success rate over 99.4\% on the 2x2x2 Rubik’s Cube, this method demonstrates the potential to address sparse-reward challenges without complex sampling or search methods.

Despite these promising results, our approach currently has some limitations. We focused on the 2x2x2 Rubik’s Cube to test feasibility within a manageable state space. Scaling to the 3x3x3 cube would require a much larger neural network to estimate costs across a more complex space, posing a key challenge for future work. Furthermore, while the 2x2x2 cube provided an initial testbed, broader testing across different sparse-reward environments will be essential to assess the generalizability and practicality of this method. Applying our approach to various domains could demonstrate its robustness for other sparse-reward tasks where rewards are infrequent or difficult to access.

In summary, our approach offers a new method for solving sparse-reward problems by optimizing policies from fully scrambled states, without relying on search or solved-state sampling. These results lay a foundation for developing more flexible and scalable methods, though further testing on complex puzzles and varied environments will be important to confirm its wider applicability.




\bibliographystyle{unsrt}

\phantomsection
\addcontentsline{toc}{section}{\refname}
\bibliography{template.bib}

\end{document}